\documentclass{article}

\usepackage{nips15submit_e} 

\usepackage{times}

\usepackage{graphicx} 
\usepackage{subfigure} 

\usepackage{booktabs}

\usepackage[numbers]{natbib}

\usepackage{algorithm}
\usepackage{algorithmic}

\usepackage{hyperref}
\usepackage{url}

\usepackage{amsmath}

\usepackage[skip=-10pt]{caption}
\author{
Matthieu Courbariaux \\
\'{E}cole Polytechnique de Montr\'{e}al \\
\texttt{matthieu.courbariaux@polymtl.ca} \\
\And
Yoshua Bengio \\
Universit\'{e} de Montr\'{e}al, CIFAR Senior Fellow \\
\texttt{yoshua.bengio@gmail.com} \\
\And
Jean-Pierre David \\
\'{E}cole Polytechnique de Montr\'{e}al \\
\texttt{jean-pierre.david@polymtl.ca} \\
}

\def\citet{\citep}

\nipsfinalcopy 

\title{BinaryConnect: Training Deep Neural Networks with binary weights during propagations}

\begin{document}

\maketitle

\begin{abstract}

Deep Neural Networks (DNN) have achieved state-of-the-art results in a wide range of tasks,
with the best results obtained with large training sets and large models. In the past, GPUs enabled these
breakthroughs because of their greater computational speed. In the future, faster computation
at both training and test time is likely to be crucial for further progress and for
consumer applications on low-power devices. As a result, there is much interest in research and development 
of dedicated hardware for Deep Learning (DL).
Binary weights, i.e., weights which are constrained to only two possible values (e.g. -1 or 1), 
would bring great benefits to specialized DL hardware by replacing many multiply-accumulate operations by
simple accumulations, as multipliers are the most space and power-hungry components of the digital implementation
of neural networks.
We introduce BinaryConnect, 
a method which consists in training a DNN with binary weights during the forward and backward propagations,
while retaining precision of the stored weights in which gradients are accumulated.
Like other dropout schemes, we show that BinaryConnect acts as regularizer and
we obtain near state-of-the-art results with BinaryConnect on the permutation-invariant MNIST, CIFAR-10 and SVHN.

\end{abstract}

\section{Introduction}

Deep Neural Networks (DNN) have substantially pushed the state-of-the-art in a wide range of tasks,
especially in speech recognition~\citep{Hinton-et-al-2012,Sainath-et-al-ICASSP2013} 
and computer vision, notably object recognition from images~\citep{Krizhevsky-2012-small,Szegedy-et-al-arxiv2014}.
More recently, deep learning is making important strides in natural language processing, especially
statistical machine translation~\citep{Devlin-et-al-ACL2014,Sutskever-et-al-NIPS2014,Bahdanau-et-al-ICLR2015-small}.
Interestingly, one of the key factors that enabled this major progress has been the advent of
Graphics Processing Units (GPUs), with speed-ups on the order of 10 to 30-fold, starting with~\citet{RainaICML09-small},
and similar improvements with distributed training~\citep{Bengio-nnlm2003,Dean-et-al-NIPS2012}.
Indeed, the ability to train larger models on more data has enabled the kind of breakthroughs
observed in the last few years. Today, researchers and developers designing new deep learning algorithms
and applications often find themselves limited by computational capability. This along, with the drive to put deep learning
systems on low-power devices (unlike GPUs) is greatly increasing the interest in research and development of 
specialized hardware for deep networks~\citep{Kim-et-al-2009, Chen-et-al-ACM2014, Chen-et-al-IEEE2014}.

Most of the computation performed during training and application of deep networks regards the multiplication
of a real-valued weight by a real-valued activation (in the recognition or forward propagation phase
of the back-propagation algorithm) or gradient (in the backward propagation phase of the
back-propagation algorithm). This paper proposes an approach called BinaryConnect to eliminate the need for these multiplications
by forcing the weights used in these forward and backward propagations to be binary, i.e. constrained
to only two values (not necessarily 0 and 1). We show that state-of-the-art results can be achieved
with BinaryConnect on the permutation-invariant MNIST, CIFAR-10 and SVHN.

What makes this workable are two ingredients:
\begin{enumerate}
\item Sufficient precision is necessary to accumulate and
average a large number of stochastic gradients, but noisy weights (and we can view discretization
into a small number of values as a form of noise, especially if we make this discretization
stochastic) are quite compatible with Stochastic Gradient Descent (SGD), the main type
of optimization algorithm for deep learning.
SGD explores the space of parameters by making small and noisy steps and that noise is
{\em averaged out} by the stochastic gradient contributions  accumulated in each weight. 
Therefore, it is important to keep sufficient resolution for these accumulators, which at first sight
suggests that high precision is absolutely required. \citet{Muller-et-al-2015} and \citet{Gupta-et-al-ICML2015}
show that randomized or stochastic rounding can be used to provide unbiased discretization.
\citet{Muller-et-al-2015} have shown that SGD requires weights with a precision of at least 6 to 8 bits
and \citet{Courbariaux-et-al-ICLR2015workshop} successfully train DNNs with 12 bits dynamic fixed-point
computation.
Besides, the estimated precision of the brain synapses varies between 6 and 12 bits~\citep{Bartol-et-al-bioRxiv2015}.
\item Noisy weights actually provide a form of regularization which can help to
generalize better, as previously shown with variational weight noise~\citep{Graves2011-variational},
Dropout~\citep{Srivastava-master-small,Srivastava14} and DropConnect~\citep{Wan+al-ICML2013-small}, which add noise to the
activations or to the weights. For instance, DropConnect \citep{Wan+al-ICML2013-small}, which is closest
to BinaryConnect, is a very efficient regularizer that randomly substitutes half of the weights with zeros during propagations.
What these previous works show is that {\em only the expected value
of the weight needs to have high precision}, and that noise can actually be beneficial.
\end{enumerate}

The main contributions of this article are the following.
\begin{itemize}
    
    
    \item We introduce BinaryConnect, 
    a method which consists in training a DNN with binary weights during the forward and backward propagations (Section \ref{sec:binary}).
    
    \item We show that BinaryConnect is a regularizer
    and we obtain near state-of-the-art results on the permutation-invariant MNIST, CIFAR-10 and SVHN (Section \ref{sec:exp}).


    \item We make the code for BinaryConnect available
    \footnote{\url{https://github.com/MatthieuCourbariaux/BinaryConnect}}.
  
\end{itemize}


\section{BinaryConnect}
\label{sec:binary}

In this section we give a more detailed view of BinaryConnect, considering which two values to choose, how to 
discretize, how to train and how to perform inference.

\subsection{$+1$ or $-1$}
\label{subsec:values}


Applying a DNN mainly consists in convolutions and matrix multiplications.
The key arithmetic operation of DL is thus the multiply-accumulate operation.
Artificial neurons are basically multiply-accumulators computing weighted sums of their inputs.

BinaryConnect constraints the weights to either $+1$ or $-1$ during propagations.
As a result, many multiply-accumulate operations are replaced by simple additions (and subtractions).
This is a huge gain, as fixed-point adders are much less expensive both in terms of
area and energy than fixed-point multiply-accumulators \citep{David-et-al-2007}.


\subsection{Deterministic vs stochastic binarization}

The binarization operation transforms the real-valued weights into the two possible values.
A very straightforward binarization operation would be based on the sign function:
\begin{equation}
    w_b = \left\{ \begin{array}{ll}
            +1 & \mbox{if $w \geq 0$},\\
            -1 & \mbox{otherwise}.\end{array} \right.
\end{equation}
Where $w_b$ is the binarized weight and $w$ the real-valued weight. Although this is
a deterministic operation, averaging this discretization over the many input weights
of a hidden unit could compensate for the loss of information. An alternative that
allows a finer and more correct averaging process to take place is to binarize
stochastically:
\begin{align}
\label{eq:sampled-wb}
    w_b = \left\{ \begin{array}{ll}
            +1 & \mbox{with probability $p = \sigma(w)$},\\
            -1 & \mbox{with probability $1-p$}.\end{array} \right. 
\end{align}
where $\sigma$ is the {\em ``hard sigmoid''} function:
\begin{equation}
    \sigma(x) = {\rm clip}(\frac{x+1}{2},0,1) = \max(0,\min(1,\frac{x+1}{2}))
\end{equation}
We use such a hard sigmoid rather than the soft version
because it is far less computationally expensive (both in software and
specialized hardware implementations) and yielded excellent results in our experiments.
It is similar to the ``hard tanh'' non-linearity introduced by~\citet{Collobert04}.
It is also piece-wise linear and corresponds to a bounded form of the rectifier~\citep{Glorot+al-AI-2011-small}.

\subsection{Propagations vs updates}
\label{subsec:prop}

Let us consider the different steps of back-propagation with SGD udpates and whether it makes
sense, or not, to discretize the weights, at each of these steps.
\begin{enumerate}
    \item Given the DNN input, compute the unit activations layer by layer, leading to the top layer which is the output of the DNN, given its input. This step is referred as the forward propagation.
    \item Given the DNN target, compute the training objective's gradient w.r.t. each layer's activations,
    starting from the top layer and going down layer by layer until the first hidden layer. 
    This step is referred to as the backward propagation or backward phase of back-propagation.
    \item Compute the gradient w.r.t. each layer's parameters
    and then update the parameters using their computed gradients and their previous values. 
    This step is referred to as the parameter update.
\end{enumerate}

\begin{algorithm}[H]
\begin{algorithmic}
    \REQUIRE a minibatch of (inputs, targets),
    previous parameters $w_{t-1}$ (weights) and $b_{t-1}$ (biases), 
    and learning rate $\eta$.
    \ENSURE updated parameters $w_t$ and $b_t$.  
    \STATE {\bf 1. Forward propagation:}
    \STATE $w_b \leftarrow {\rm binarize}(w_{t-1})$
    \STATE For $k=1$ to $L$, compute $a_k$ knowing $a_{k-1}$, $w_b$ and $b_{t-1}$
    \STATE {\bf 2. Backward propagation:}
    \STATE Initialize output layer's activations gradient $\frac{\partial C}{\partial a_L}$
    \STATE For $k=L$ to $2$, compute $\frac{\partial C}{\partial a_{k-1}}$
          knowing $\frac{\partial C}{\partial a_k}$ and $w_b$
    \STATE {\bf 3. Parameter update:}
    \STATE Compute $\frac{\partial C}{\partial w_b}$ and $\frac{\partial C}{db_{t-1}}$ 
        knowing $\frac{\partial C}{\partial a_k}$ and $a_{k-1}$
    \STATE $w_t \leftarrow {\rm clip}(w_{t-1} - \eta \frac{\partial C}{\partial w_b})$
    \STATE $b_t \leftarrow b_{t-1} - \eta \frac{\partial C}{\partial b_{t-1}}$
\end{algorithmic}
\caption{SGD training with BinaryConnect. $C$ is the cost function for minibatch and the functions binarize($w$) and clip($w$) specify how to binarize and clip weights. $L$ is the number of layers.
}
\label{alg:train}
\end{algorithm}

A key point to understand with BinaryConnect is that we only binarize the weights during the forward and backward propagations
(steps 1 and 2) but not during the parameter update (step 3), as illustrated in Algorithm \ref{alg:train}.
Keeping good precision weights during the updates is necessary for SGD to work at all.
These parameter changes are tiny by virtue of being obtained
by gradient descent, i.e., SGD performs a large number of almost infinitesimal changes in the direction that most
improves the training objective (plus noise). 
One way to picture all this is to hypothesize that what matters most at the end of training is the sign of 
the weights, $w^*$, but that in order to figure it out, we perform a lot of small changes to a continuous-valued
quantity $w$, and only at the end consider its sign: 
\begin{equation}
   w^* = {\rm sign}(\sum_t g_t)
\end{equation}
where $g_t$ is a noisy estimator of $\frac{\partial C(f(x_t,w_{t-1},b_{t-1}),y_t)}{\partial w_{t-1}}$,
where $C(f(x_t,w_{t-1},b_{t-1}),y_t)$ is the value of the objective function on (input,target) example $(x_t,y_t)$, 
when $w_{t-1}$ are the previous weights and $w^*$ is its final discretized value of the weights.

Another way to conceive of this discretization is as a form of corruption, and hence as a regularizer, and our 
empirical results confirm this hypothesis. In addition, {\em we can make the discretization errors on different weights
approximately cancel each other while keeping a lot of precision by randomizing the discretization
appropriately.} We propose a form of randomized discretization that {\em preserves the expected
value of the discretized weight.} 

Hence, at training time, BinaryConnect randomly picks one of two values for each weight, for each
minibatch, for both the forward and backward propagation phases of backprop. However, the SGD update
is accumulated in a real-valued variable storing the parameter.

An interesting analogy to understand BinaryConnect is the DropConnect algorithm~\citep{Wan+al-ICML2013-small}.
Just like BinaryConnect, DropConnect only injects noise to the weights during the propagations. Whereas
DropConnect's noise is added Bernouilli noise, BinaryConnect's noise is a binary sampling process. In both cases
the corrupted value has as expected value the clean original value.

\subsection{Clipping}

Since the binarization operation is not influenced by variations of the real-valued weights $w$
when its magnitude is beyond the binary values $\pm 1$, and since it is a common practice
to bound weights (usually the weight vector) in order to regularize them, we have chosen to
clip the real-valued weights within the $[-1,1]$ interval right after the weight updates, 
as per Algorithm \ref{alg:train}.
The real-valued weights would otherwise grow very large without any impact on the binary weights.

\subsection{A few more tricks}

\begin{table}[h]
\begin{center}
\begin{tabular}{@{}lll@{}}
\toprule
Optimization      & No learning rate scaling & Learning rate scaling \\ \midrule
SGD               &                          & 11.45\%               \\
Nesterov momentum & 15.65\%                  & 11.30\%               \\
ADAM              & 12.81\%                  & \textbf{10.47\%}      \\ \bottomrule
\end{tabular}
\end{center}
\caption{Test error rates of a (small) CNN trained on CIFAR-10 
    depending on optimization method and on whether the learning rate is scaled
    with the weights initialization coefficients from \citep{GlorotAISTATS2010-small}.}
\label{tab:tricks}
\end{table}

We use Batch Normalization (BN) \citep{Ioffe+Szegedy-2015} in all of our experiments,
not only because it accelerates the training by reducing internal covariate shift,
but also because it reduces the overall impact of the weights scale.
Moreover, we use the ADAM learning rule \citep{kingma2014adam} in all of our CNN experiments.
Last but not least, we scale the weights learning rates 
respectively with the weights initialization coefficients from \citep{GlorotAISTATS2010-small} when optimizing with ADAM,
and with the squares of those coefficients when optimizing with SGD or Nesterov momentum \citep{Nesterov83}.
Table \ref{tab:tricks} illustrates the effectiveness of those tricks.


\subsection{Test-Time Inference}
\label{subsec:inference}

Up to now we have introduced different ways of {\em training} a DNN with on-the-fly weight binarization.
What are reasonable ways of using such a trained network, i.e., performing test-time inference on new examples?
We have considered three reasonable alternatives:
\begin{enumerate}
\item Use the resulting binary weights $w_b$ (this makes most sense with the deterministic form of BinaryConnect).
\item Use the real-valued weights $w$, i.e., the binarization only helps to achieve faster training but
not faster test-time performance.
\item In the stochastic case, many different networks can be sampled by sampling a $w_b$ for each weight
according to Eq.~\ref{eq:sampled-wb}. The ensemble output of these networks can then be obtained by averaging
the outputs from individual networks. 
\end{enumerate}
We use the first method with the deterministic form of BinaryConnect.
As for the stochastic form of BinaryConnect, 
we focused on the training advantage and used the second method in the experiments, i.e.,
test-time inference using the real-valued weights. This follows the practice of Dropout methods, where
at test-time the ``noise'' is removed.

\begin{table}[h]
\begin{center}
\begin{tabular}{@{}llll@{}}
\toprule
Method                                        & MNIST             & CIFAR-10        & SVHN            \\ \midrule
No regularizer                                & 1.30 $\pm$ 0.04\% & 10.64\%         & 2.44\%          \\
BinaryConnect (det.)                          & 1.29 $\pm$ 0.08\% & 9.90\%          & 2.30\%          \\
BinaryConnect (stoch.)                        & 1.18 $\pm$ 0.04\% & \textbf{8.27\%} & 2.15\%          \\
50\% Dropout                                  & 1.01 $\pm$ 0.04\% &                 &                 \\ \midrule
Maxout Networks \citep{Goodfeli-et-al-TR2013} & 0.94\%            & 11.68\%         & 2.47\%          \\
Deep L2-SVM \citep{Tang-wkshp-2013}           & \textbf{0.87\%}   &                 &                 \\
Network in Network \citep{Lin-et-al-2013}     &                   & 10.41\%         & 2.35\%          \\
DropConnect \citep{Wan+al-ICML2013-small}     &                   &                 & 1.94\%          \\
Deeply-Supervised Nets \citep{Lee-et-al-2014} &                   & 9.78\%          & \textbf{1.92\%} \\ \bottomrule
\end{tabular}
\end{center}
\caption{
   Test error rates of DNNs trained on the MNIST (no convolution and no unsupervised pretraining), 
   CIFAR-10 (no data augmentation) and SVHN, depending on the method.
   We see that in spite of using only a single bit per weight during propagation, performance
   is not worse than ordinary (no regularizer) DNNs, it is actually better, especially with the
   stochastic version, suggesting that BinaryConnect acts as a regularizer. }
\label{tab:res}
\end{table}


\begin{figure}[ht]
\begin{center}
\centerline{\includegraphics[width=\textwidth]{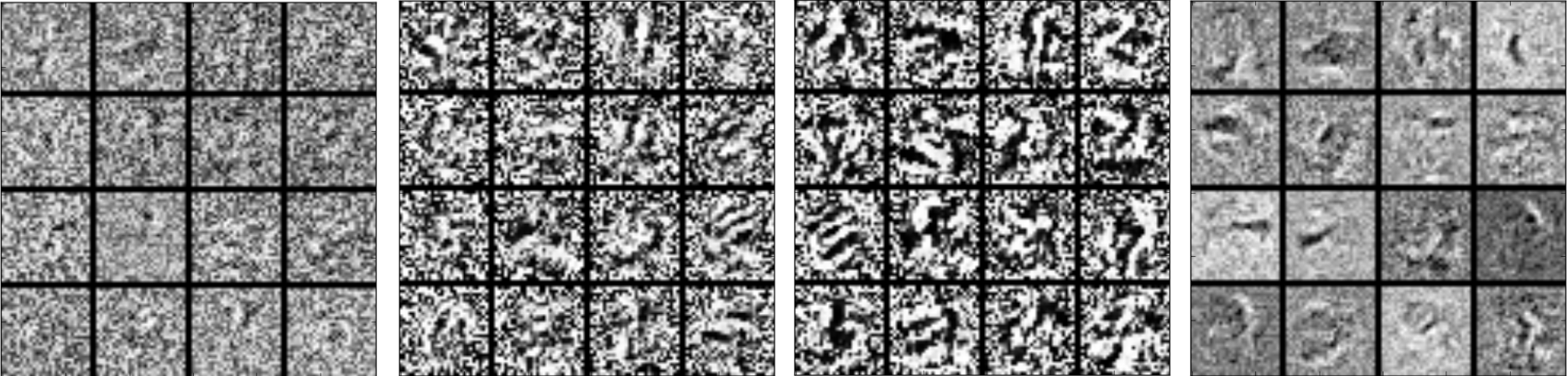}}
\end{center}
\caption{
Features of the first layer of an MLP trained on MNIST depending on the regularizer.
From left to right: no regularizer, deterministic BinaryConnect, stochastic BinaryConnect and Dropout.
}
\label{fig:features}
\end{figure}

\begin{figure}[ht]
\begin{center}
\centerline{\includegraphics[width=.75\textwidth]{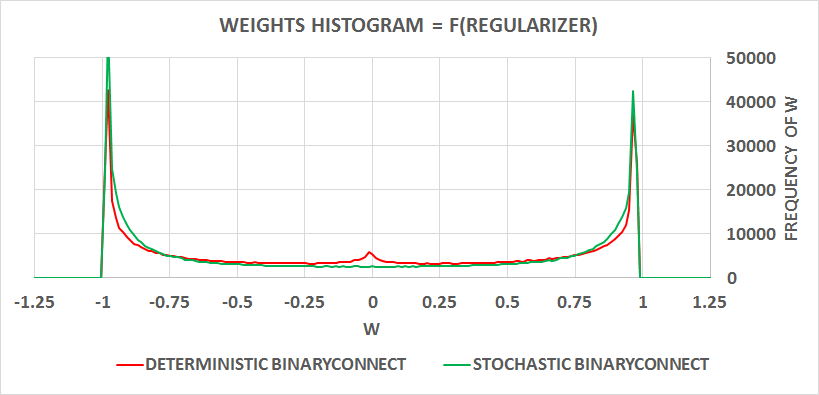}}
\end{center}
\caption{
Histogram of the weights of the first layer of an MLP trained on MNIST depending on the regularizer.
In both cases, it seems that the weights are trying to become deterministic to reduce the training error.
It also seems that some of the weights of deterministic BinaryConnect are stuck around 0, hesitating between $-1$ and $1$.
}
\label{fig:histograms}
\end{figure}


\begin{figure}[ht]
\begin{center}
\centerline{\includegraphics[width=.9\textwidth]{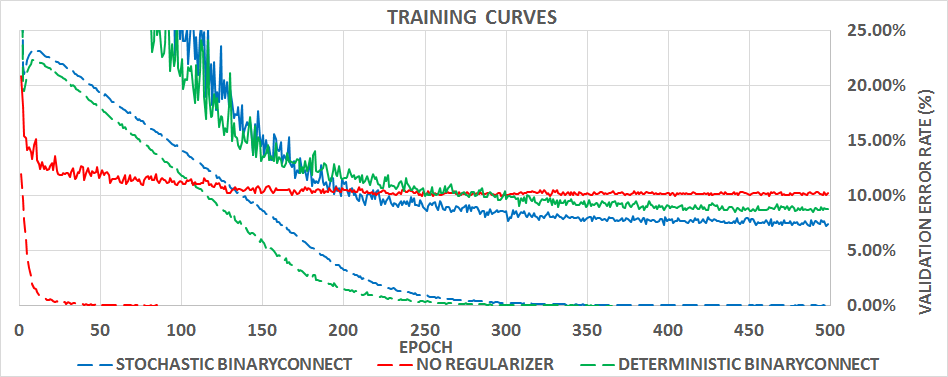}}
\end{center}
\caption{
Training curves of a CNN on CIFAR-10 depending on the regularizer.
The dotted lines represent the training costs (square hinge losses)
and the continuous lines the corresponding validation error rates.
Both versions of BinaryConnect significantly augment the training cost, slow down the training
and lower the validation error rate, which is what we would expect from a Dropout scheme.  
}
\label{fig:curves}
\end{figure}



\section{Benchmark results}
\label{sec:exp}

In this section, 
we show that BinaryConnect acts as regularizer and
we obtain near state-of-the-art results with BinaryConnect on the permutation-invariant MNIST, CIFAR-10 and SVHN.

\subsection{Permutation-invariant MNIST}

MNIST is a benchmark image classification dataset \citep{LeCun+98}.
It consists in a training set of 60000 and a test set of 10000 28 $\times$ 28 gray-scale images representing digits ranging from 0 to 9.
Permutation-invariance means that the model must be unaware of the image (2-D) structure of the data (in other words, CNNs are forbidden).
Besides, we do not use any data-augmentation, preprocessing or unsupervised pretraining.
The MLP we train on MNIST consists in 3 hidden layers of 1024 Rectifier Linear Units (ReLU)
\citep{Nair-2010-small,Glorot+al-AI-2011-small,Krizhevsky-2012-small}
and a L2-SVM output layer (L2-SVM has been shown to perform 
better than Softmax on several classification benchmarks \citep{Tang-wkshp-2013,Lee-et-al-2014}).
The square hinge loss is minimized with SGD without momentum.
We use an exponentially decaying learning rate.
We use Batch Normalization with a minibatch of size 200 to speed up the training.
As typically done, we use the last 10000 samples of the training set as a validation set for early stopping and model selection.
We report the test error rate associated with the best validation error rate after 1000 epochs
(we do not retrain on the validation set).
We repeat each experiment 6 times with different initializations.
The results are in Table \ref{tab:res}.
They suggest that the stochastic version of BinaryConnect can be considered a regularizer, 
although a slightly less powerful one than Dropout, in this context.

\subsection{CIFAR-10}

CIFAR-10 is a benchmark image classification dataset.
It consists in a training set of 50000 and a test set of 10000 32 $\times$ 32 color images representing 
airplanes, automobiles, birds, cats, deers, dogs, frogs, horses, ships and trucks.
We preprocess the data using global contrast normalization and ZCA whitening.
We do not use any data-augmentation (which can really be a game changer for this dataset \citep{Graham-2014}).
The architecture of our CNN is:
\begin{equation}
    (2 \times 128C3)-MP2-(2 \times 256C3)-MP2-(2 \times 512C3)-MP2-(2 \times 1024FC)-10SVM
\end{equation}
Where $C3$ is a $3\times3$ ReLU convolution layer, $MP2$ is a $2\times2$ max-pooling layer, 
$FC$ a fully connected layer, and SVM a L2-SVM output layer.
This architecture is greatly inspired from VGG \citep{Simonyan2015}.
The square hinge loss is minimized with ADAM.
We use an exponentially decaying learning rate.
We use Batch Normalization with a minibatch of size 50 to speed up the training.
We use the last 5000 samples of the training set as a validation set.
We report the test error rate associated with the best validation error rate after 500 training epochs
(we do not retrain on the validation set).
The results are in Table \ref{tab:res} and Figure \ref{fig:curves}.

\subsection{SVHN}

SVHN is a benchmark image classification dataset.
It consists in a training set of 604K and a test set of 26K 32 $\times$ 32 color images representing 
digits ranging from 0 to 9.
We follow the same procedure that we used for CIFAR-10, with a few notable exceptions:
we use half the number of hidden units and we train for 200 epochs instead of 500 (because SVHN is quite a big dataset).
The results are in Table \ref{tab:res}.

\section{Related works}
\label{sec:related}

Training DNNs with binary weights has been the subject of very recent works
\citep{Soudry-et-al-NIPS2014-small,Cheng-et-al-2015,hwang-et-al-2014,kim-et-al-2014}.
Even though we share the same objective, our approaches are quite different.
\citep{Soudry-et-al-NIPS2014-small,Cheng-et-al-2015} do not train their DNN with Backpropagation (BP) 
but with a variant called Expectation Backpropagation (EBP).
EBP is based on Expectation Propagation (EP) \citep{Minka-2001-small},
which is a variational Bayes method used to do inference in probabilistic graphical models.
Let us compare their method to ours:
\begin{itemize}
    \item It optimizes the weights posterior distribution (which is not binary).
    In this regard, our method is quite similar as we keep a real-valued version of the weights.
    \item It binarizes both the neurons outputs and weights, 
    which is more hardware friendly than just binarizing the weights.
    \item It yields a good classification accuracy for fully connected networks (on MNIST)
    but not (yet) for ConvNets.
\end{itemize}

\citet{hwang-et-al-2014,kim-et-al-2014} {\em retrain} neural networks 
with {\em ternary} weights during forward and backward propagations, i.e.:
\begin{itemize}
    \item They train a neural network with high-precision,
    \item After training, they ternarize the weights to three possible values $-H$, $0$ and $+H$
        and adjust $H$ to minimize the output error,
    \item And eventually, they retrain with ternary weights during propagations and high-precision weights during updates.
\end{itemize}
By comparison, we {\em train all the way} with {\em binary} weights during propagations, i.e., 
our training procedure could be implemented with efficient specialized hardware 
avoiding the forward and backward propagations multiplications,
which amounts to about $2/3$ of the multiplications (cf. Algorithm \ref{alg:train}).

\section{Conclusion and future works}



We have introduced a novel binarization scheme for weights during forward and backward propagations
called BinaryConnect.
We have shown that it is possible to train DNNs with BinaryConnect 
on the permutation invariant MNIST, CIFAR-10 and SVHN datasets
and achieve nearly state-of-the-art results.
The impact of such a method on specialized hardware implementations of deep
networks could be major, by removing the need for about 2/3 of the multiplications,
and thus potentially allowing to speed-up by a factor of 3 at training time.
With the deterministic version of BinaryConnect the impact at test time could
be even more important, getting rid of the multiplications altogether  
and reducing by a factor of at least 16 (from 16 bits single-float
precision to single bit precision) the memory requirement of deep networks,
which has an impact on the memory to computation bandwidth and on the size of
the models that can be run.
Future works should extend those results to other models and datasets,
and explore getting rid of the multiplications altogether during training, by
removing their need from the weight update computation.

 
\section{Acknowledgments}

We thank the reviewers for their many constructive comments.
We also thank Roland Memisevic for helpful discussions.
We thank the developers of Theano \citep{bergstra+al:2010-scipy,Bastien-Theano-2012},
a Python library which allowed us to easily develop a fast and optimized code for GPU.
We also thank the developers of Pylearn2 \citep{pylearn2_arxiv_2013} 
and Lasagne \citep{dieleman-et-al-2015},
two Deep Learning libraries built on the top of Theano.
We are also grateful for funding from NSERC, the Canada Research Chairs, Compute Canada, Nuance Foundation, and CIFAR.

{\small
\bibliography{strings,strings-shorter,aigaion,ml,discrete_weights}
\bibliographystyle{unsrt}
}
\end{document}